\documentclass[runningheads]{llncs}
\usepackage{graphicx}

\usepackage{tikz}
\usepackage{comment}
\usepackage{color}
\usepackage{multirow}
\usepackage{graphicx}
\usepackage{booktabs}
\usepackage{algorithm}
\usepackage{algorithmic}
\usepackage{url}
\usepackage{threeparttable}
\usepackage{float}
\usepackage[accsupp]{axessibility}  

\begin{document}
\pagestyle{headings}
\mainmatter
\def\ECCVSubNumber{4048}  

\title{JCDNet: Joint of Common and Definite phases Network for Weakly Supervised Temporal
Action Localization} 

\titlerunning{JCDNet}
%
\author{Yifu Liu\inst{1} \and
Xiaoxia Li\inst{1} \and
Zhiling Luo\inst{1} \and Wei Zhou\inst{1}}
\authorrunning{Liu et al.}
%
\institute{Alibaba DAMO Academy\inst{1}\\
\email{zhencang.lyf@alibaba-inc.com}\\
}
\maketitle

\begin{abstract}
Weakly-supervised temporal action localization aims to localize action instances in untrimmed videos with only video-level supervision. 
We witness that different actions record common phases, {\it e.g.}, the run-up in the HighJump and LongJump.
These different actions are defined as conjoint actions, whose rest parts are definite phases, {\it e.g.}, leaping over the bar in a HighJump.
Compared with the common phases, the definite phases are more easily localized in existing researches. 
Most of them formulate this task as a Multiple Instance Learning paradigm, in which the common phases are tended to be confused with the background, and affect the localization completeness of the conjoint actions.
To tackle this challenge, we propose a Joint of Common and Definite phases Network (JCDNet) by improving feature discriminability of the conjoint actions. 
Specifically, we design a Class-Aware Discriminative module to enhance the contribution of the common phases in classification by the guidance of the coarse definite-phase features.
Besides, we introduce a temporal attention module to learn robust action-ness scores via modeling temporal dependencies, distinguishing the common phases from the background.
Extensive experiments on three datasets (THUMOS14, ActivityNetv1.2, and a conjoint-action subset) demonstrate that JCDNet achieves competitive performance against the state-of-the-art methods.

\end{abstract}

\section{Introduction}

With only video-level annotation during the training stage to localize action instances in an untrimmed video, weakly-supervised temporal action localization (WTAL) attracts the attention of researchers.
The unclear boundaries in the temporal sequences and large temporal spans in the long untrimmed video bring austere challenges to the WTAL tasks.
Most previous works \cite{lee2020background,islam2021hybrid,lee2021WTAL-Uncertainty} focus on separating foreground actions from background.
In addition, some works \cite{liu2019completeness,liu2021acsnet} try to divide the background phase into two parts: normal background and context, for precise localization. 

\begin{figure}[!ht]
	\centering
	\includegraphics[width=0.85\linewidth]{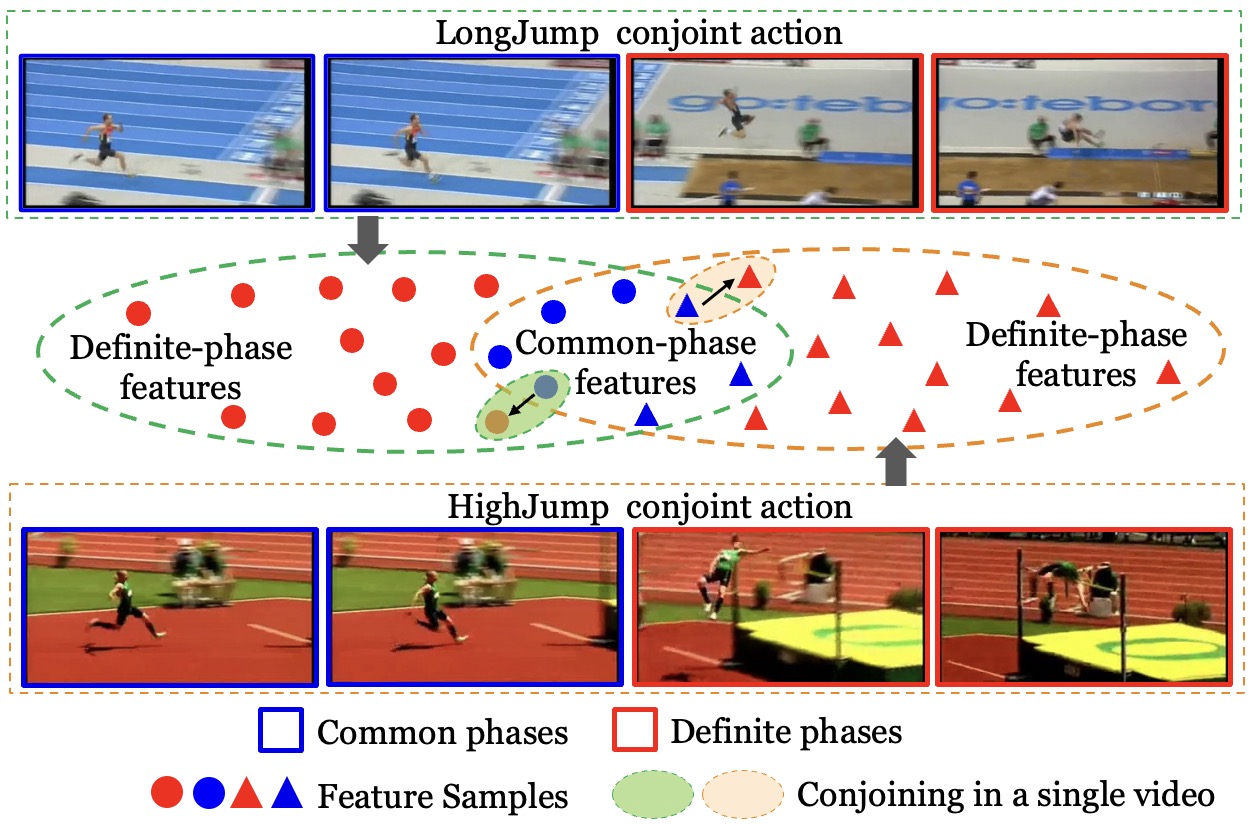}
	\caption{
	Both LongJump conjoint action and HighJump conjoint action are composed of common phases in blue boxes and definite phases in red boxes.
	The green and red dashed circles are conjoint action features of LongJump and HighJump respectively.
	The red markers are definite-phase features, while the blue markers are misclassified common-phase features in the MIL paradigm.
	Our method utilizes the definite-phase features in a given video to guide the common phases to be identified correctly.
	}
	\label{Fig1}            
\end{figure}

This work introduces a novel perspective for the WTAL task, learning the localization features of an action with the help of other actions. 
Generally speaking, some overlap phases are observed in different actions.
For example, the phase demonstrating run-up, exists in not only HighJump but also LongJump.
As illustrated in Figure \ref{Fig1}, the complete LongJump and HighJump start with run-up (as common phases in blue boxes), but end with different identified actions (as definite phases in red boxes).
To simplify our discussion, we name the overlap parts as \textit{common phases} and the identified as the \textit{definite phases}.
Besides, those actions containing common and definite phases are called \textbf{conjoint actions}.
Note that neither common nor definite phases are annotated in the dataset. We even don't know whether an action is conjoint or not.
Therefore, to make use of this perspective in WTAL, we need an implicit model to 1) learn the representation of common/definite phases from the videos, and 2)  localize them in a given video.

When the action instances are conjoint actions, the existing methods could not classify the common phases as the interest action.
Since most existing WTAL methods are formulated as multiple instance learning (MIL) tasks, the models in the MIL paradigm tend to aggregate the discriminative features, namely the {\it definite-phase} features above (red markers in Figure \ref{Fig1}).
Specifically, the generated snippet-level class scores, known as Temporal Class Activation Sequences (T-CAS) \cite{paul2018w}, will be aggregated to video-level classification scores. 
Final localization relies on a threshold of the class activations.
Considering the action classification and localization are non-simultaneous, and the MIL tasks aim at minimizing the video-level classification loss, the network will eventually predict the definite phases with high scores while ignoring other parts of the action (common phases). 
The common phases can easily be misclassified and cause performance degradation.

As analyzed above, we found that conjoining common phases and definite phases
is the key to predict more accurate boundaries in the conjoint action localization (CAL). 
In light of this, we propose a \textbf{J}oint of \textbf{C}ommon and \textbf{D}efinite phases Network (\textbf{JCD}Net) for precise localization in CAL.


We contend that the definite-phase features offer a powerful inductive bias to help involve the common phases.
It is impossible to directly model the common-phase features with definite-phase features due to the absence of snippet-level annotations.
To tackle this challenge, we introduce the Class-Aware Discriminative (CAD) module, a process to approximately model the feature representation of conjoint action (color masks shown in Figure \ref{Fig1}).
By leveraging the class representation of T-CAS, the snippet-level features are aggregated into new coarse definite-phase features in a given video (red markers under color masks in Figure 1). 
Then we use the coarse definite-phase features to adaptively reweight each snippet representation. This helps enhance awareness of the common-phase snippets in classification from candidate snippets
(blue markers under color masks in Figure \ref{Fig1}).

In fact, separating the background and foreground is still a hard work because the common-phase actions frequently co-occur with confusing background. 
A Temporal-Enhanced Attention (TEA) module is introduced to capture features dependencies in the temporal dimension, enhancing feature representations and learning robust action-ness scores for background suppression.
Extensive experiments on two benchmarks, {\it i.e}, THUMOS14 dataset \cite{THUMOS14} and ActivityNet1.2 dataset \cite{caba2015activitynet}, show that the proposed method achieves competitive performance against the state-of-the-art methods.
Spotlighting on the CAL task, we present a conjoint-action subset, which contains all eleven action categories in THUMOS14 that can be defined as conjoint actions.
Several experiments on this subset are implemented and show that our method outperforms the state-of-the-art methods in the CAL task.

To summarize, our contribution is three-fold:
\begin{itemize}
\item 
We provide a new perspective: {\it conjoint action}, which contains common phases and definite phases.
And we find it hard to localize the conjoint action precisely.
To tackle this challenge,
we propose a Joint of Common and Definite phases Network (JCDNet) to conjoin the common and definite phases to an entire action and distinguish foreground from background.
\item In JCDNet, we propose a Class-Aware Discriminative module to model the coarse definite-phase features to enhance awareness of the common-phase snippets in classification.
Moreover, a Temporal-Enhanced Attention module is designed to capture temporal dependencies and suppress background activities.
\item The proposed method achieves competitive performance against the state-of-the-art methods on two public benchmarks and a conjoint-action subset. And we will release the code at:
\url{https://drive.google.com/file/d/1qGqIBJuqfOAkXFX9mKT1Nim5cmVFvL_/view?usp=sharing}.
\end{itemize}

\section{Related Work}
\textbf{Fully-supervised Temporal Action Localization.} 
Fully-supervised TAL utilizes frame-level annotations of all action instances during training. 
Many existing works \cite{zhao2017temporal,chao2018rethinking,lin2018bsn,lin2020fast,heilbron2017scc}
apply a two-stage pipeline, i.e., generating proposals by
sliding window and classifying proposals into different
actions or background activity. 
Besides, several works \cite{sstad_buch_bmvc17,lin2017single,long2019gaussian,zhao2020bottom} in a one-stage pipeline directly predict action category and temporal boundaries followed by some post-processing techniques.
Although these fully-supervised methods have achieved good performance, they severely depend on frame-level annotations.
\\
\textbf{Weakly-supervised Temporal Action Localization.} 
WTAL only requires video-level supervision and has drawn extensive attention. Most existing WTAL methods are formulated as the MIL framework \cite{paul2018w,lee2020background,islam2021hybrid,ji2021weakly}.
W-TALC \cite{paul2018w} and A2CL-PT \cite{min2020adversarial} employed deep metric learning to make the distance of the features in the same category closer than in different categories.
To solve the problem that backgrounds are easily misclassified,
Nguyen {\it et al.} \cite{nguyen2019weakly} and BaS-Net \cite{lee2020background} introduced an auxiliary background class for modeling background features.
Ham-Net \cite{islam2021hybrid} adopted a hybrid attention mechanism to model an action instance with discriminative parts.
Since context frequently co-occur with the action,
ACSNet \cite{liu2021acsnet} explicitly took context into account for accurate action localization.
UM-Net \cite{lee2021WTAL-Uncertainty} proposed to formulate background frames as out-of-distribution samples and model uncertainty with feature magnitudes.
CoLA \cite{zhang2021cola} introduced a Snippet Contrast Loss to refine the hard snippet representation for precise temporal boundaries.
Also, several works \cite{hong2021cross,ji2021weakly} attempted to explore the cross-modal feature fusion and improve the feature discrimination capability.

To our best knowledge, this work is the first one that introduces a new perspective of the given dataset, {\it i.e.}, conjoint action, and takes the conjoint action localization (CAL) as the key to the WTAL task.
The proposed JCDNet focuses on identifying the common phases via modeling the coarse definite-phase features and distinguishing them from the background.

\section{Methodology}

In this section, we define our tasks, introduce the architecture of our JCDNet, and finally present details of training and inference, respectively.

\subsection{Problem Formulation}
\label{3.1}
Assume that we are provided with a set of videos, each video containing several action instances.
Only video-level label $y$ is available, where $y \in \mathbb{R}^{C+1}$ is a $C+1$ dimension multi-hot vector,
$C+1$ corresponds to $C$ class of actions and one background class.
Our goal is to localize all action instances in an untrimmed video.
Each action instance is denoted as
$\left(t_{s}, t_{e}, \psi, c\right)$, where $t_{s}$ and $t_{e}$ denote the start and end timestamps of an action, $c$ is the predicted action class, and $\psi$ represents the confidence score for this action proposal. 

\begin{figure}[t]
	\centering
	\includegraphics[width=1.0\linewidth]{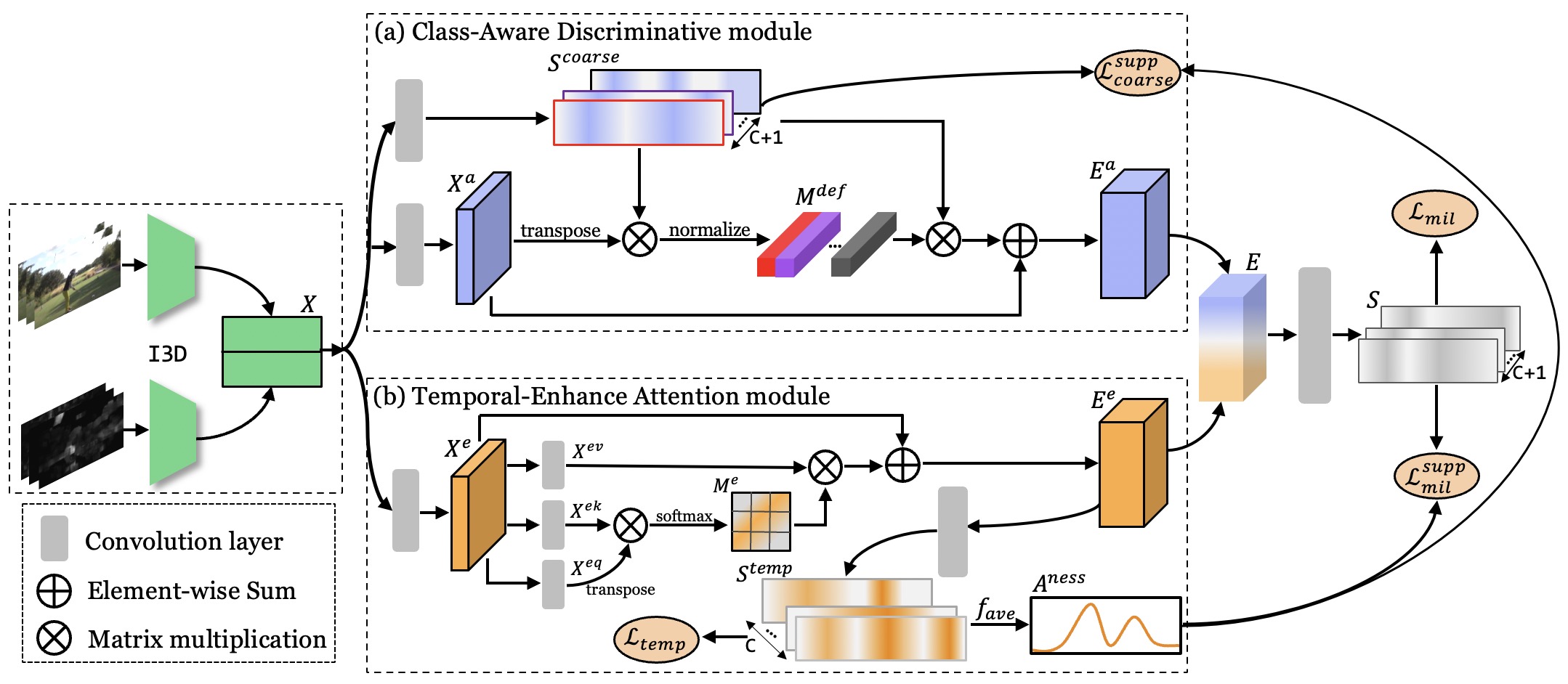}
	\caption{
	Overview of the proposed JCDNet. 
	It mainly consists of two parts: (a) Class-Aware Discriminative module and (b) Temporal-Enhanced Attention module.
	}
	\label{Fig2}            
\end{figure}

\subsection{Pipeline} 
\label{3.2}
\textbf{Feature Extraction:}
Following conventions, we sample a fixed number of $T$ snippets from each video due to the variation of video length. 
And then we extract the features of RGB $X^{RGB} \in \mathbb{R}^{F \times T}$ and 
optical flow $X^{flow} \in \mathbb{R}^{F \times T}$ from pre-trained extractor, {\it i.e.,} I3D \cite{carreira2017quo}. The concatenated features $X \in \mathbb{R}^{2 F \times T}$ of both RGB and optical flow features are fed into our model.
\\
\textbf{Structure Overview:}
Figure \ref{Fig2} shows the whole pipeline of the proposed JCDNet.
The snippet-level features $X$ are fed into CAD and TEA modules, which are class-level and temporal attention mechanisms, respectively.
In the CAD module, we leverage the snippet-level features and T-CAS to calculate the coarse definite-phase feature representations in a given video. 
Then a global view is provided to guide the common phases from the candidate snippets to be classified correctly.
In the TEA module, a self-attention unit is adopted for capturing enhanced features along the temporal dimension. 
Then the enhanced features are fused with the output of the CAD module and are utilized to generate action-ness scores for further background suppression.

\subsection{Class-Aware Discriminative Module}
\label{3.3}
\textbf{Distinguish the coarse definite-phase features.} 
In the CAD module, our goal is to utilize definite-phase features for improving the awareness of common-phase features in classification. 
Intuitively, we need to model a coarse definite-phase features due to the absence of snippet-level annotations.
As mentioned above, models tend to aggregate the definite-phase features under a MIL supervision manner. In the MIL paradigms, the intermediate T-CAS presents the class possibility distributions of each snippet.
So we can leverage the T-CAS to 
collect coarse definite-phase features in a given video.

\begin{figure}[t]
	\centering
	\includegraphics[width=0.7\linewidth]{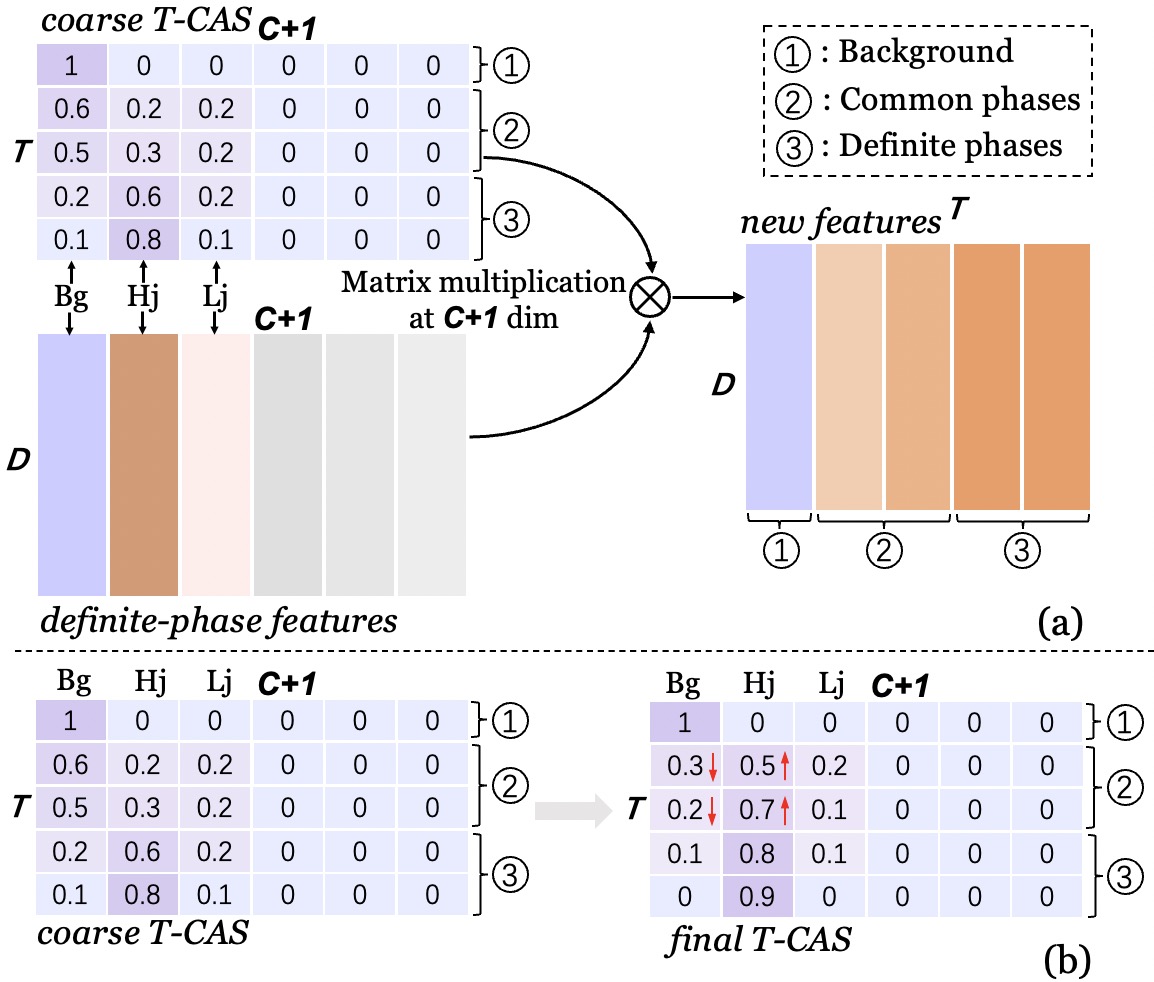}
	\caption{
	(a) With the help of coarse T-CAS, each candidate snippet finds the similarity to the coarse definite-phase features.
	(b) With the supervised process, the common phases, misclassified with high scores in the coarse T-CAS (left in b), are correctly classified in the final T-CAS. Bg/Hj/Lj represent background, HighJump and LongJump classes, respectively.
	}
	\label{Fig4}            
\end{figure}

As shown in Figure \ref{Fig2}, given the input feature $X \in \mathbb{R}^{2 F \times T}$, a convolution layer is adopted to generate the coarse T-CAS $S^{coarse} \in \mathbb{R}^{(C+1) \times T}$.
Meanwhile, another convolution layer is adopted to obtain the more task-specific features $X^{a} \in \mathbb{R}^{D \times T}$.
Then we perform a matrix multiplication and normalization between $S^{coarse}$ and transposed $X^{a}$ to calculate coarse definite-phase features $M^{def} \in \mathbb{R}^{(C+1) \times D}$.
The process is formulated as follows:
\begin{equation}
M_{c}^{def}=\frac{\sum_{t=1}^{T} S_{t, c}^{coarse} \cdot X_{t}^{a}}{\sum_{t=1}^{T} S_{t, c}^{coarse}}
\end{equation}

where $S_{t, c}^{coarse}$ represents the probability of the snippet $t$ belonging to the $c$-th class, and $M_{c}^{ def},X_{t}^{a} \in \mathbb{R}^{1 \times D}$.
In this process, snippets with higher class activation contribute more weights for calculating definite features of one class in a given video.
\\
\textbf{Re-classify common phases from candidate snippets.}
Then we get the coarse definite-phase features, which can enhance the model's discriminative ability for each candidate snippet.
Further, the definite-phase features provide a global view to enhance awareness of common-phase snippets in classification and achieve the conjoining with definite phases.
As shown in Figure \ref{Fig4}(a), the global view is provided by matrix multiplication between the $S^{coarse}$ and transposed $M^{def}$. In detail,
we perform a matrix multiplication between the $S^{coarse}$ and transposed $M^{def}$, and then multiply the result by a scale parameter $\alpha$ and perform an element-wise sum operation with $X^{a}$ to obtain the fine features $E^{a} \in \mathbb{R}^{D \times T}$:
\begin{equation}
E^{a}=\alpha \sum_{c=1}^{C+1} S_{c}^{coarse} \cdot M_{c}^{def} + X^{a}
\end{equation}

where $\alpha$ is initialized as 0 and learned by stochastic gradient descent. 
Even if common-phase snippets are misclassified as background, the network will provide the inductive bias for them by exploiting the definite-phase features reweighting the representation of the $S^{coarse}$.
Intuitively speaking, $M^{def}$ helps reclassify the candidate snippets as actions by finding their similarities to the definite snippets.
Hence $E^{a}$ reflects that each snippet can adaptively conjoin definite-phase features of different classes.

To better understand the process, we visualize the supervised process of snippets' classification with T-CAS, as shown in Figure \ref{Fig4}(b). After the supervision, the misclassified candidate snippets will be correctly classified with the help of the fine features.

\subsection{Temporal-Enhanced Attention Module}
\label{3.4}
Untrimmed videos usually contain a lot of background snippets, easily confused with the common-phase snippets.
A general method to suppress background activity \cite{islam2021hybrid,zhang2021cola}
is to predict action-ness scores of each snippet through simple temporal convolution layers.
Due to the lack of computing the relationships between different snippets in a video, this approach ignores the temporal contextual dependencies and affects the representation of the action-ness scores.

Therefore, we first introduce a self-attention mechanism to capture contextual dependencies in temporal dimensions.
As illustrated in Figure \ref{Fig2}, given the input feature 
$X \in \mathbb{R}^{2 F \times T}$, we first feed it into a convolution layer to generate the base feature $X^{e} \in \mathbb{R}^{D \times T}$.
Then, we feed $X^{e}$ into the followed convolution to generate two new feature $X^{eq}$ and $X^{ek}$, respectively, where 
$\{X^{eq}, X^{ek}\} \in \mathbb{R}^{D \times T}$.
The temporal attention map $M^{e} \in \mathbb{R}^{T \times T}$ is formulated as follows:
\begin{equation}
M_{t_{i}t_{j}}^{e}=\frac{\exp \left(X_{t_{i}}^{eq} \cdot X_{ t_{j}}^{ek}\right)}{\sum_{t_{i}=1}^{T} \exp \left(X_{ t_{i}}^{eq} \cdot X_{ t_{j}}^{ek}\right)}
\end{equation}

where $M_{t_{i}t_{j}}^{e}$ represents the $t_{i}$ snippet's impact on the $t_{j}$ snippet. 
To this extent, snippets with similar feature representations contribute to higher temporal correlation.
Then $X^{e}$ is fed into a convolution layer to generate a new feature $X^{ev} \in \mathbb{R}^{D \times T}$. We perform a matrix multiplication between $X^{ev}$ and the transpose of $M^{e}$.
Finally, we multiply it by a scale parameter $\beta$ and perform a element-wise sum operation with the feature $X^{e}$ to obtain the enhanced feature $E^{e} \in \mathbb{R}^{D \times T}$:
\begin{equation}
E^{e}=\beta \sum_{t=1}^{T} M_{t}^{e} \cdot X_{t}^{ev} + X^{e}
\end{equation}

where $\beta$ gradually learns from 0 by stochastic gradient descent.
After the attention unit, the enhanced feature $E^{e}$ at each snippet can be aware of the temporal context according to the attention map. 

The enhanced feature $E^{e}$ is fed into a class-specific convolution layer to obtain the T-CAS $S^{temp} \in \mathbb{R}^{C \times T}$ with the temporal contextual dependencies.
Note that $S^{temp}$ represents the temporal activation information of C-class actions without background.
Then the average pooling operation $f_{ave}$ is applied for acquiring robust action-ness scores $A^{ness} \in \mathbb{R}^{T \times 1}$. The process is formulated as follows:

\begin{equation}
A^{ness}= {Sigmoid}\left(f_{ave}\left(S^{temp}\right)\right)
\end{equation}

Note, $A^{ness}$ inherits the property of $S^{temp}$, and separates the backgrounds and actions. Thus, $A^{ness}$ presents the overall distribution of foreground scores.
Further, we adopt the enhanced feature $E^{e}$ with $E^{a}$ for feature fusion and the robust action-ness scores $A^{ness}$ are used for background suppression. 
More discussions are explained in Section \ref{3.5}.

\subsection{Optimizing Process}
\label{3.5}
We concatenate the class-aware feature $E^{a}$ and the temporal-enhanced feature $E^{e}$ to form fusion features $E\in \mathbb{R}^{2D \times T}$ and feed it into the final classifier to produce the final T-CAS $S \in \mathbb{R}^{(C+1) \times T}$.


In order to enhance the suppression of background, we design two suppression losses
$\mathcal{L}_{mil}^{supp}$ and $\mathcal{L}_{coarse}^{supp}$
, to supervise the whole process and the construction of the definite-phase features.
The specific processing is as follows:
we apply the action-ness scores $A^{ness}$ to suppress the background in T-CAS $S^{coarse}$ and $S$.
The suppressed T-CAS $\overline{S}^{coarse}$ and $\overline{S}$ are formulated as follows:
\begin{equation}
\begin{aligned}
\overline{S}^{coarse} &= A^{ness} \odot S^{coarse}\\
\overline{S} &= A^{ness} \odot S 
\end{aligned}
\end{equation}

where $\odot$ is element-wise multiplication. $A^{ness}$ is adopted to supervise the $S^{coarse}$ and $S$, which help supervise the process of obtaining the coarse definite-phase features and the entire conjoint-action features, respectively. Then $S^{coarse}$ and $S$ rescale their gradients for each parameter. 

We apply the top-k MIL loss \cite{paul2018w} on the $\overline{S}$ and $\overline{S}^{coarse}$, denoted as $\mathcal{L}_{mil}^{supp}$ and $\mathcal{L}_{coarse}^{supp}$. 
Since both $\mathcal{L}_{mil}^{supp}$ and $\mathcal{L}_{coarse}^{supp}$ have the same form, we only present the process of $\mathcal{L}_{mil}^{supp}$ below, and replace $\overline{S}$ with $\overline{S}^{coarse}$ to get the process of $\mathcal{L}_{coarse}^{supp}$.
Following \cite{islam2021hybrid,ji2021weakly}, we aggregate the top-k values among all snippet-level scores in Eq.\ref{eq7}, and apply a softmax function to obtain video-level class scores $\overline{p}_{j}$:

\begin{equation}
\overline{v}(j)=\max _{l \subset\{1,2, \ldots, T\} \atop|l|=k} \frac{1}{k} \sum_{i \in l} \overline{S}_{i}(j)
\label{eq7}
\end{equation}
\begin{equation}
\overline{p}(j)=\frac{\exp \left(\overline{v}(j)\right)}{\sum_{j^{\prime}=1}^{C+1} \exp \left(\overline{v}(j^{\prime})\right)}
\end{equation}

where $j=1,2, \ldots, C+1$. 
The loss $\mathcal{L}_{mil}^{supp}$ is defined as the cross-entropy loss between the video label and the prediction $\overline{p}(j)$,
\begin{equation}
\mathcal{L}_{mil}^{supp}=-\sum_{j=1}^{C+1} y^{f}(j) \log \left(\overline{p}(j)\right)
\end{equation}

where $y^{f}(j)$ contains only foreground activities, {\it i.e.,} the background class $y^{f}(C+1)=0$ for suppressing the background activity. 
And the final suppressed MIL loss can be formulated as,
\begin{equation}
\mathcal{L}_{supp} = \mathcal{L}_{mil}^{supp} + \lambda_{0} \mathcal{L}_{coarse}^{supp}
\end{equation}


Besides, we apply the co-activity similarity loss
$\mathcal{L}_{cas}$\footnote{The co-activity similarity are widely used in WTAL methods and is not the main contribution in this work. More details can refer to \cite{paul2018w}.} \cite{paul2018w} on the enhanced feature $E^{a}$ and the suppressed coarse T-CAS $\overline{S}^{coarse}$ to model better definite-phase feature representation.
Following the work \cite{nguyen2018weakly}, we utilize a L1-norm loss $\mathcal{L}_{norm}$ to make foreground weights more polarized, denoted as $\mathcal{L}_{norm}= \sum_{i=1}^{T}\left|A^{ness}_{i}\right|$, where $\left|\cdot\right|$ is a L1-norm function.
In addition, we introduce a guidance loss $\mathcal{L}_{guide}$ to make the distribution of action-ness scores $A^{ness}$ opposite to the background class probability in $S$, {\it i.e.,} $\mathcal{L}_{guide}= \sum_{i=1}^{T}\left|1-A^{ness}_{i}-s_{c+1}\right|$, where $s_{c+1}$ is the last column in the T-CAS $S$ that represents the background class probability of each snippet.

Also, we apply the same top-k MIL loss on the final T-CAS $S$ and the TEA module's $S^{temp}$, 
\begin{equation}
\mathcal{L}_{MIL} = \mathcal{L}_{mil} + \lambda_{1} \mathcal{L}_{temp}
\end{equation}

Finally, we train JCDNet with the following loss function:
\begin{equation}
\begin{aligned}
\mathcal{L} &=
\mathcal{L}_{MIL}+
\mathcal{L}_{supp}+
\lambda_{2} \mathcal{L}_{cas}+
\lambda_{3} \mathcal{L}_{norm} +
\lambda_{4} \mathcal{L}_{guide}
\end{aligned}
\end{equation}

where $\lambda_{0}$ and $\lambda_{1}$ mentioned above and $\lambda_{2}$, $\lambda_{3}$ and $\lambda_{4}$ here are hyper-parameters to balance the contribution of each loss function. 

\subsection{Temporal Action Localization}
\label{3.6}
During inference, we first calculate the video-level class score based on $\overline{S}$ and discard classes with a video-level class score less than a particular threshold (set to 0.2 in our experiments).
For the remaining classes, we threshold the action-ness scores $A^{ness}$ to discard the background snippets and obtain the class-agnostic action proposals by selecting the continuous segments of the remaining snippets. 
Since a candidate action locations is a four-tuple:$\left(t_{s}, t_{e}, \psi, c\right)$, we utilize the suppressed T-CAS $\overline{S}$ to calculate the class-specific score $\psi$ for each action proposal following the process of \cite{islam2021hybrid}.
Besides, we apply different thresholds for obtaining
action proposals and remove the overlapping segments with non-maximum suppression.

\section{Experiments}

\subsection{Experimental Settings}
\textbf{Dataset.} 
We evaluate our approach on three dataset: two public benchmark datasets, {\it i.e.}, THUMOS14 \cite{THUMOS14} and ActivityNet1.2 \cite{caba2015activitynet}, and a conjoint-action subset.
\textbf{THUMOS14} provides 20 action categories annotations, including 200 untrimmed validation videos and 213 test videos. 
Following the previous work, we use validation videos for training and test videos for test.
\textbf{ActivityNet1.2} provides temporal annotations for 100 action categories, including 4,819 untrimmed videos for training and 2,383 untrimmed videos for testing.
It contains around 1.5 activity segments (10 times sparser than THUMOS14) and 36\% background per video.
\textbf{Conjoint-action subset}
contains eleven action categories in THUMOS14, where each action can find the common phases in other actions belonging to this subset (More details in \textbf{Section \ref{sec4.3}}).
\\
\textbf{Evaluation metrics.}
We evaluate our method with mean Average Precision (mAP) under several different intersection of union (IoU) thresholds, which are the standard evaluation metric for temporal action localization \cite{paul2018w}. 
Moreover, we utilize the official evaluation code of ActivityNet \cite{caba2015activitynet} to calculate the evaluation metrics.
\\
\textbf{Implementation details.}
For feature extraction, we employ I3D network \cite{carreira2017quo} pre-trained on Kinetics \cite{kay2017kinetics} to extract both RGB and optical flow features, where the optical flow is created using the TV-L1 algorithm \cite{wedel2009improved}.
We sample continuous non-overlapping 16 frames from video as a snippet, then concatenate both streams extracted features to obtain 2048-dimensional snippet-level features.
During training, we randomly sample 500 snippets for THUMOS14 and 60 snippets for ActivityNet1.2 to fix the input segments, and we take all the snippets during evaluation.

We use the Adam optimizer with learning rate 1e-4 and weight decay
rate 1e-3, and train 100 epochs for THUMOS14 and 20 epochs for ActivityNet1.2.
$\lambda_{0}$, $\lambda_{1}$, $\lambda_{2}$, $\lambda_{3}$, $\lambda_{4}$ are set 0.8, 0.7, 0.9, 0.8, 0.8, respectively, to obtain the best performance for both datasets.
We sample 20 videos in a batch during training, including 3 pairs of videos, each containing the same categorical tag for co-activity similarity loss $\mathcal{L}_{cas}$.
For action localization, we utilize a set of thresholds from 0.05 to 0.95 with a step of 0.06, and perform non-maximum suppression (NMS) with threshold 0.7 to remove highly overlapping proposals.
Note, we utilize the open-source model of other methods, released by the authors, to report the results for fair comparisons.

\begin{table*}[htbp]
\centering
\caption{
Comparisons with the state-of-the-art methods on THUMOS14. $^{\dagger}$ indicates the use of additional labels, \textit{i.e.,} action frequency or human pose.}
 \resizebox{0.95\linewidth}{!}{
\begin{tabular}{c|l|ccccccc|cc}
\toprule
\multirow{2}{*}{Supervision} & \multicolumn{1}{c|}{\multirow{2}{*}{Method}} & \multicolumn{7}{c|}{mAP@IoU}            & \multicolumn{2}{c}{AVG} \\  
                             & \multicolumn{1}{c|}{}                    & 0.3 & 0.4 & 0.5 & 0.6 & 0.7 & 0.8 & 0.9 & 0.5:0.9    & 0.3:0.9    \\ \hline\hline
\multirow{4}{*}{Fully}       & S-CNN \cite{shou2016temporal}  &36.3     &28.7     &19.0     &10.3     &5.3     &-     &-     &-            &-            \\
& SSN \cite{zhao2017temporal}                 &50.6     &40.8     &29.1     &-     &-     &-     &-     &-            &-            \\
                             & BSN \cite{lin2018bsn}                       &53.5     &45.0     &36.9     &28.4     &20.0     &-     &-     &-            &-  
                             \\ 
                             & G-TAD \cite{xu2020g}                        &54.5     &47.6     &40.2     &30.8     &23.4     &-     &-     &-            &- 
                             \\ \hline
\multirow{3}{*}{Weakly$^{\dagger}$}      & STAR \cite{xu2019segregated}                &48.7     &34.7     &23.0     &-     &-     &-     &-     &-            &-            \\
& Nguyen {\it et al.} \cite{nguyen2019weakly}   &49.1     &38.4     &27.5     &17.3     &8.6     &3.2     &0.5     &11.4            &20.7            \\
                             & SF-Net \cite{ma2020sf}                       &53.2     &40.7     &29.3     &18.4     &9.6     &-     &-     &-   &-            \\ \hline
\multirow{8}{*}{Weakly}      & BaS-Net \cite{lee2020background} &44.6     &36.0     &27.0     &18.6     &10.4     &3.3     &0.4     &11.9   &20.0   \\
& A2CL-PT \cite{min2020adversarial}                                       &48.1     &39.0     &30.1     &19.2     &10.6     &4.8     &1.0     &13.14            &21.8 \\
& TSCN \cite{zhai2020two}                                       &47.8     &37.7     &28.7     &19.4     &10.2     &3.9     &0.7     &12.6            &21.2 \\
& ACSNet \cite{liu2021acsnet}                                   &51.4     &42.7     &32.4     &22.0     &11.7     &-     &-     &-            &- \\
                             & HAM-Net \cite{islam2021hybrid}               &50.3     &41.1     &31.0     &20.7     &11.4     &4.4     &0.7     &13.64   &22.8
                             \\ 
 
                             & UM \cite{lee2021WTAL-Uncertainty}               &52.3     &43.4     &33.7     &22.9     &12.1     &3.9     &0.4     &14.6    &24.1\\
& CoLA \cite{zhang2021cola} &52.1     &43.1     &34.3     &23.5     &13.1     &3.7     &0.6     &17.0   &25.8
                             \\                             
                             & CO$_{2}$-Net \cite{hong2021cross}               &\textbf{54.5}     &\textbf{45.7}     &38.3     &\textbf{26.4}     &13.4     &6.9     &2.0     &17.4            & \textbf{26.7}\\
& Ours            &52.6     &44.9     &\textbf{38.4}     &25.8    &\textbf{14.2}     &\textbf{7.6 }    &\textbf{2.1}           &\textbf{17.6} &26.6\\
                             \bottomrule
\end{tabular}}
\label{tab1}
\end{table*}

\subsection{Comparisons with State-of-the-art Methods}
\textbf{Experiments on THUMOS14.} Table \ref{tab1} demonstrates the performance comparisons between our proposed method and the state-of-the-art fully/weakly supervised TAL methods on the THUMOS14 dataset.
Reconsidering our method and spotlighting on the completeness of the entire action, it is more applicable for precise localization.
Therefore, we care more about the mAP under high IoU thresholds, and we report mAP scores under the IoU thresholds settings from 0.3 to 0.9, and present two kinds of average mAP (0.5:0.9 and 0.3:0.9).
Overall, our method performs well against existing weakly supervised methods, especially in terms of the average mAP from IoU 0.5 to 0.9.
Specifically, JCDNet outperforms best at the IoU threshold of 0.5, 0.7, 0.8, and 0.9, which proves the effectiveness of our method.
\\
\textbf{Experiments on ActivityNet.}
The performance on the ActivityNet1.2 dataset is demonstrated in Table \ref{tab5}.
As shown, consistent with the results on THUMOS14, our method obtains comparable performance with other state-of-the-art WTAL methods, following SSN \cite{zhao2017temporal}  with a small gap, less than 0.2\%.

\begin{table*}[htbp]
\centering
\caption{
Comparisons with the state-of-the-art methods on ActivityNet1.2.
AVG(0.5:0.95) denotes the average mAP at IoU thresholds from 0.5 to 0.95 with 0.05 increment.}
\renewcommand\tabcolsep{7.0pt} 
\begin{threeparttable}
\begin{tabular}{c|l|ccc|c}
\toprule
\multirow{2}{*}{Supervision} & \multicolumn{1}{c|}{\multirow{2}{*}{Method}} & \multicolumn{3}{c|}{mAP@IoU} & \multicolumn{1}{c}{AVG} 
\\ 
& \multicolumn{1}{c|}{}    & 0.5 & 0.75 & 0.95     & 0.5:0.95    \\ \hline\hline
                             
\multirow{1}{*}{Fully-Supervised}       & SSN \cite{zhao2017temporal}                  &41.3     &27.0     &6.1     &26.6             \\\hline
\multirow{8}{*}{Weakly-Supervised} 
& TSCN \cite{zhai2020two}      &37.6     &23.7     &5.7    &23.6 \\
& BaS-Net \cite{lee2020background}  &38.5     &24.2     &5.6       &24.3   \\
& HAM-Net \cite{islam2021hybrid}     &41.0     &24.8     &5.3            &25.1\\ 
& UM \cite{lee2021WTAL-Uncertainty}           &41.2     &25.6     &6.0                    &25.9 \\
& ACSNet \cite{liu2021acsnet}                                   &40.1     &26.1     &\textbf{6.8}            &26.0 \\
& CoLA \cite{zhang2021cola}     &42.7    &25.7     &5.8            &26.1 \\
& CO$_{2}$-Net \cite{hong2021cross}              &\textbf{43.3}     &26.3     &5.2             & 26.4\\
& Ours            &43.0     &\textbf{26.5}     &5.5        &\textbf{26.4}\\
                             \bottomrule
\end{tabular}
\end{threeparttable}
\label{tab5}
\end{table*}

\subsection{Performance of Conjoint Action Localization}
\label{sec4.3}
Regarding of conjoint action we proposed in the introduction, our goal is to differentiate conjoint actions with similar common phases.
In order to prove our conjecture, we construct all conjoint actions in THUMOS14 as 
an independent conjoint-action subset to perform the experiments.
The subset contains five sets with eleven classes of different actions, where the actions within each set contain similar common phases but different definite phases, so these actions can be defined as conjoint actions.
These sets are ``BaseballPitch" \& ``CricketBowling", ``HighJump" \& ``LongJump", ``HammerThrow" \& ``Shotput" \& ``ThrowDiscus", ``JavelinThrow" \& ``PoleVault", and ``CliffDiving" \& ``Diving".
Table \ref{tab4} summarizes performance comparisons on this subset.
There is no doubt that we achieve state-of-the-art results, which validates the effectiveness of our method from a quantitative perspective.
Note that we do not adopt additional annotations on the subset for fair comparison.

\begin{figure}[tb]
	\centering
	\includegraphics[width=1.0\linewidth]{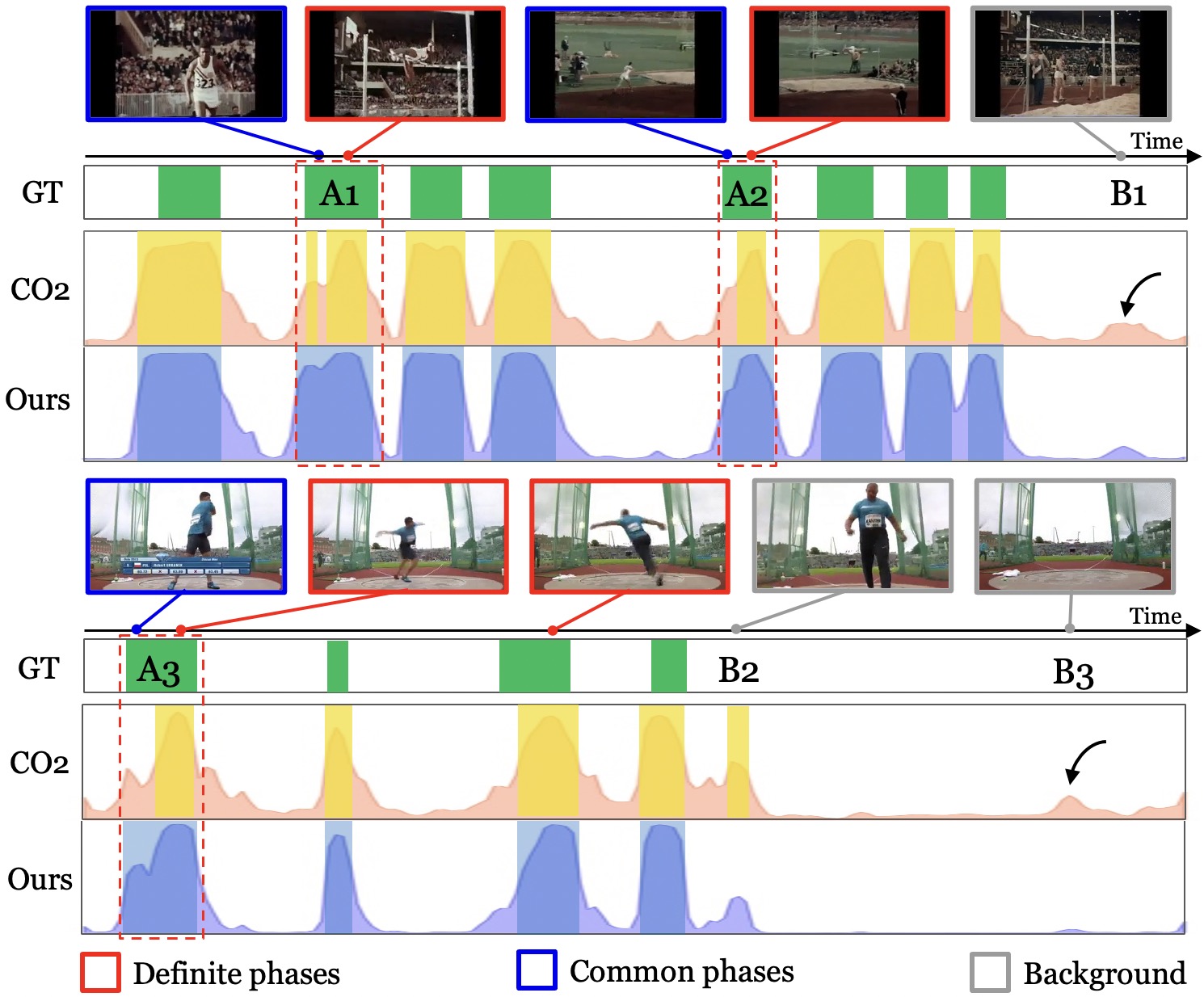}
	\caption{
	The illustration of the action localization results, predicted by the CO$_{2}$-Net and our method, on ``HighJump" (top) and ``ThrowDiscus" (bottom). The horizontal and vertical axes are time and the intensity of action scores, respectively. For clarity, frames with red bounding boxes refer to definite phases, those in blue refer to common phases, and those in gray refer to backgrounds.
	}
	\label{Fig3}  
\end{figure}

To further illustrate the effectiveness of our method, we visualize the comparison with the state-of-the-art method CO$_{2}$-Net \cite{hong2021cross} on some conjoint actions, \textit{i.e.}, ``HighJump" top and ``ThrowDiscus" bottom in Figure \ref{Fig3}. 
As shown in Figure \ref{Fig3}, action instances of A$_{1}$, A$_{2}$ and A$_{3}$ are representative samples to show the impressive ability of our model, which provide higher confidence scores for each action snippet and conjoin the common and definite phases to an entire action.
With the absence of global awareness of the definite-phase features, CO$_{2}$-Net divides action A$_{1}$ into two actions and ignores the common-phase parts of action A$_{2}$ and A$_{3}$.
Moreover, our model performs well in suppressing the background, \textit{i.e.,} activities of B$_{1}$, B$_{2}$, and B$_{3}$ in Figure \ref{Fig3}. 
In particular, CO$_{2}$-Net incorrectly
predicts B$_{1}$ as an action. 
Note, all background activities get lower action-ness scores.

\begin{table*}[htb]
\centering
\caption{
Comparisons with the state-of-the-art methods on the conjoin-action subset.}
\renewcommand\tabcolsep{6.5pt} 
\begin{threeparttable}
\begin{tabular}{l|ccccccc|c}
\toprule
\multicolumn{1}{c|}{\multirow{2}{*}{Method}} & \multicolumn{7}{c|}{mAP@IoU} & \multicolumn{1}{c}{AVG} \\  
 \multicolumn{1}{c|}{}& 0.3 & 0.4 & 0.5 & 0.6 & 0.7 & 0.8 & 0.9   & 0.3:0.9    \\ \hline\hline


 UM                &46.8     &37.7     &26.9     &16.2     &7.5     &2.9     &0.2        &19.8\\
  CoLA          &48.2     &38.6     &29.5     &18.7     &9.9     &2.7     &0.5      &21.2\\ 
 CO$_{2}$-Net      &67.2     &59.0     &49.8     &34.9     &19.7     &9.7     &2.5    & 34.7\\
 Ours              &\textbf{67.7}    &\textbf{60.8}     &\textbf{52.9}     &\textbf{36.2}     &\textbf{21.2}     &\textbf{11.2}    &\textbf{2.9} &\textbf{36.1}\\
\bottomrule
\end{tabular}
\end{threeparttable}
\label{tab4}
\end{table*}

\subsection{Ablation Study}
In this section, we perform multiple ablation studies on THUMOS14 to analyze the performance contribution of each component in JCDNet. 
The baseline is similar to the Main pipeline in the UM \cite{lee2021WTAL-Uncertainty}.
The extracted feature $X$ is fed into the last convolution layer as our baseline (``Exp 1" in Table \ref{tab3}), where merely multiple-instance learning loss $\mathcal{L}_{mil}$ is employed. 
\\
\textbf{Effect of class-aware discriminative module.} 
To demonstrate the effectiveness of the CAD module, we drop the TEA module and the supervision with action-ness scores.
We only incorporate the CAD module achieves 19.8\% average of mAP, which brings
6.5\% improvement compared with the baseline (``Exp 2" in Table \ref{tab3}).
Our CAD module can enhance the awareness of common-phase snippets in classification by the guidance of the coarse definite-phase features.
To further verify the effect of definite-phase features, we remove the MIL supervision used to generate the coarse T-CAS so that the definite-phase features can not be exploited explicitly (``Exp 3" in Table \ref{tab3}). 
The comparison results in Table \ref{tab3} show that employing the whole CAD module outperforms the result of ``Exp 3" by 4.9\%.
\\
\textbf{Effect of temporal-enhanced attention module.} 
To evaluate the performance of the TEA module, we apply a set of experiments.
Dropping the CAD module, we only utilize the TEA module outperforms the baseline by 5.8\% gain (``Exp 4" in Table \ref{tab3}). 
Compared with no suppression parts (``Exp 2" in Table \ref{tab3}), $\mathcal{L}_{mil}^{supp}$ and  $\mathcal{L}_{coarse}^{supp}$ can improve 3.6\% and 2.3\%, respectively (``Exp 5" and ``Exp 6" in Table \ref{tab3}).
When we utilize the two suppression losses together (``Exp 7" in Table \ref{tab3}), the performance further improves to 24.5\% through the mutual influence between action-ness scores and two T-CASs ($S^{coarse}$ and $S$).

\begin{table}[ht]
\centering
\caption{Ablation studies of our method in term of average mAP under multiple IoU thresholds from 0.3 to 0.9 with 0.1 increment.}
\renewcommand\tabcolsep{4.5pt} 
\begin{threeparttable}
\begin{tabular}{c|c|cc|c|cc|ccc|c}
\toprule
Exp & $\mathcal{L}_{mil}$ & CAD &n-CAD & TEA & $\mathcal{L}_{mil}^{supp}$ & $\scriptsize \mathcal{L}_{coarse}^{supp}$  & $\mathcal{L}_{norm}$ & $\mathcal{L}_{guide}$ & $\mathcal{L}_{cas}$ & AVG \\
\midrule
1       & \checkmark  &  &&&&&&& &13.3    \\
2       &\checkmark   &\checkmark      &&&&&&&&19.8  \\
3       &\checkmark   &       &\checkmark&&&&&&&14.9\\
4       &\checkmark   &&       &\checkmark&\checkmark&&&&&19.1\\
\midrule
5        &\checkmark   &\checkmark &      &\checkmark&\checkmark&&&&&23.4\\

6        &\checkmark   &\checkmark  &     &\checkmark&&\checkmark&&&&22.1\\

7        &\checkmark   &\checkmark  &     &\checkmark&\checkmark&\checkmark&&&&24.5 \\
\midrule
8        &\checkmark   &\checkmark  &     &\checkmark&\checkmark&\checkmark&\checkmark&&&25.2\\
9       &\checkmark   &\checkmark    &   &\checkmark&\checkmark&\checkmark&\checkmark&\checkmark&&26.0\\
10       &\checkmark   &\checkmark   &    &\checkmark&\checkmark&\checkmark&\checkmark&\checkmark&\checkmark&26.6\\
\bottomrule
\end{tabular}
\end{threeparttable}
\label{tab3}
\end{table}

\section{Conclusion}
We point out that conjoint action localization is critical in the WTAL task.
Then we present a novel framework JCDNet to conjoin the common phases and definite phases for localizing an entire action.
A Class-Aware Discriminative module is proposed to better identify the common-phase snippets via modelling coarse definite-phase features. 
And we utilize the Temporal-Enhanced Attention module to capture temporal dependencies and suppress the backgrounds.
The proposed method achieves outstanding performance on three datasets,
{\it i.e.,} THUMOS14, ActivityNetv1.2 and a conjoint-action subset.
Also, We believe it would be a promising research direction to introduce the conjoint action to other related tasks.

\clearpage
%

%
\bibliographystyle{splncs04}
\bibliography{egbib}
\end{document}